\documentclass[fleqn,12pt]{wlscirep}
\usepackage[utf8]{inputenc}
\usepackage[T1]{fontenc}
\title{Active 3D weaves for load-bearing and damage-resilient locomotion}

\usepackage{caption}
\usepackage{helvet}  
\makeatletter
\renewcommand{\@makecaption}[2]{%
  {\fontfamily{phv}\selectfont\footnotesize\textbf{#1.} #2\par}%
}
\makeatother 

\author[1,*]{Guowei Wayne Tu}
\author[2,3]{Evgueni T. Filipov}
\affil[1]{Mechanics and Materials of Bio-Integrated Electronics Laboratory, Department of Aerospace Engineering and Engineering Mechanics, University of Texas, Austin, TX, USA}
\affil[2]{Reconfigurable, Architected, and Regenerative Structures Laboratory, Department of Civil and Environmental Engineering, University of Michigan, Ann Arbor, MI, USA}
\affil[3]{Reconfigurable, Architected, and Regenerative Structures Laboratory, Department of Mechanical Engineering, University of Michigan, Ann Arbor, MI, USA}
\affil[*]{E-mail: guoweitu@utexas.edu}

\begin{abstract}
The craft of weaving, where different materials are interlaced, has tremendous potential for creating active and functional systems for use in soft robots, prosthetics, wearables, exoskeletons, and more. These textile-like systems are flexible and safe for human-machine interaction; however, this inherent flexibility limits their ability to carry loads, which is essential for many robotic functions. In this work, we introduce a general framework for integrating active materials into three-dimensional (3D) woven shells to create robotic structures that combine high axial stiffness for load bearing, low bending stiffness for efficient actuation, and system-level resilience for damage tolerance. These woven robots can be modularly assembled from ‘woven corners’, a fundamental unit of 3D woven structures. We use eigenvalue calculations to identify load bearing and actuation mechanisms of the 3D woven structures, and use that information to make five different robots capable of locomotion. We demonstrate that these 3D woven robots can locomote carrying loads 70 times their self-weight, and can maintain repeatable performance even after being subjected to extreme compression. This work is a pathway toward the design, manufacturing, and simulation of future 3D woven robotic systems where load bearing, high stiffness, active functional deformation, locomotion, and system-level resilience are all needed.

\end{abstract}
\begin{document}

\flushbottom
\maketitle
%
%
\thispagestyle{empty}

\section*{Introduction}\label{Sec:Intro}
Weaving is one of the oldest ways to organize slender elements into useful structures \cite{el2011basketry,martinez2023earliest}. The light weight, flexibility, and resilience have made woven structures central to daily life for centuries. Those same attributes now make woven structures attractive for robotics. Recent robotic textiles combine woven fabric-based substrates with active elements such as pneumatic fibers \cite{marshall2023mechanical,cappello2018exploiting,sanchez2021textile,chen2026knitted}, liquid crystal elastomers \cite{yang2025weaving,zhang2026weavable,silva2023active,qi2025scalable}, and shape memory alloys \cite{shin2024woven,duan2026structured,park2020novel} to produce motion while remaining soft, wearable, and safe for human contact. This combination is especially attractive for future intelligent garments and assistive devices \cite{mehr2023artificial,yun2021body,shveda2022wearable,zhang2022learning}, where actuation must be added without sacrificing comfort, portability, or body-conforming behavior \cite{williams2025openexo,pulvirenti2025resistive,kim2024skin}.

Yet, softness alone is not enough and not appropriate for many engineering tasks. Search-and-rescue devices, assistive wearables, and many other robotic systems often need to carry loads, resist collapse, and transmit forces for functional usage \cite{kim2025architected,wang2022small,carton2025bridging}. Prior studies on load-bearing soft robots have addressed this need in several complementary ways. For example, mechanical couplings have been used to create a soft robotic arm that remains compliant in bending while being up to 52 times stiffer in torsion, allowing it to continuously transmit torque for manipulation \cite{carton2025bridging}. Beaded metamaterials are naturally soft and flexible yet can deliver a large force when the cord is tightened and the beads are in contact \cite{dreier2025beaded,yan2024self}. Interlocked granular fabrics behave like fluid due to the chain mail-like architecture while can turn to a rigid solid under negative pressure where the interlocked particles are jammed \cite{wang2021structured,zhou20253d,lu2024role,yan2025rigidity}. These studies show that load bearing is an active problem in soft robotics. However, this capability has not yet been built into woven robotic textiles. Most active woven sheets remain fundamentally two-dimensional, so their motion is bending-dominated and their load transfer is typically interrupted by buckling \cite{buckner2020roboticizing,zakharov2018active,granberry2021kinetically,tajiri2026buckling}. The key opportunity is therefore to bring load-bearing modes into woven robots without losing the wearability, adaptability, and resilience that make woven textiles valuable in the first place.

\begin{figure}
\centering
\includegraphics[width=1\textwidth]{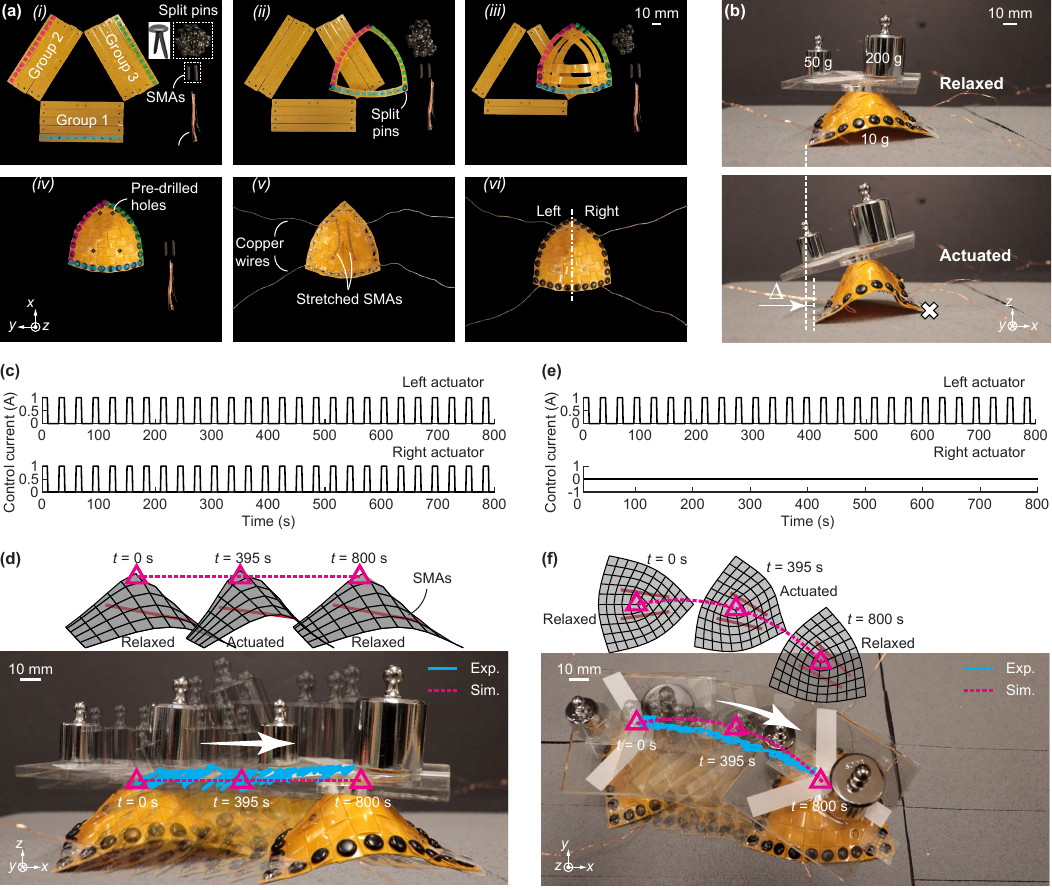}
{\caption{Fabrication and locomotion of 3D woven corners integrated with SMAs. (a) Fabrication sequence from flat ribbons to a 3D woven corner structure combined with SMAs; (b) Actuation of the woven corner structure in (a) where the SMAs are heated through a direct current; (c) and (d) are the control signal and configurations of the woven corner robot in locomotion along a straight line; (e) and (f) are the control signal and configurations of the woven corner robot in steering along a curved line. In (c) and (e), the control current signals for the left and right SMA actuators are given above and below, respectively. In (d) and (f), the images of the moving woven corner are juxtaposed with the simulated and experimental paths (blue and magenta curves); the simulated configurations are given at three discrete times ($t=$ 0, 395, 800 s), and the images are nine snapshots with the same time duration between each one.}\label{fig:Fab}}
\end{figure}

Recent work on three-dimensional woven baskets suggests a way to resolve this trade-off. Unlike flat weaves, 3D woven structures can be stiff in axial loading yet flexible in lateral bending. They carry load through stretching-dominated filament deformation, but they can also undergo large recoverable deformations through bending and local elastic buckling of filaments \cite{tu2025corner,tu2025engineering,krankel2026stiffness}. In this work, we translate that principle into robotics by using the woven corner: a topological building block that joins woven patches into a 3D form. By integrating coiled shape memory alloys into this 3D woven corner, we create woven robots that can locomote, steer, and carry heavy load through cyclic electro-thermal actuation (see Fig. \ref{fig:Fab}).

We then use the woven corner as a fundamental unit to modularly design and fabricate a family of 3D load-bearing woven robots with body-driven and limb-driven locomotion (see Fig. \ref{fig:Modular}). To describe their motion, we develop a reduced-order bar \& hinge model that predicts step sizes and locomotion paths \cite{tu2025engineering}. The model also clarifies the origin of multifunctionality in these woven systems: load bearing is governed by stiff high-order eigenmodes associated with filament stretching, whereas efficient actuation is enabled by flexible low-order eigenmodes associated with filament bending. Guided by simulations and experiments, we further show that these robots continue to locomote under loads up to 70 times their self-weight. The woven robots also maintain repeatable trajectories through 10 compression cycles, with each compression reducing them to one-quarter of their initial height. This work establishes active 3D weaves as a route toward robotic textiles that combine wearability, load bearing, programmability, and extraordinary resilience.

\section*{Integrating Active Materials into 3D Plain Weaves}\label{Sec:Snags}
In this section, we detail how we fabricate 3D woven corners, which are woven structures inspired by the topology of plain-woven baskets, and how we integrate shape memory alloys (SMAs) into woven corners as actuators to enable locomotion and steering. Then, we demonstrate how we use 3D woven corners as fundamental units to modularly design 3D robotic weaves with more complex geometries.

\subsection*{Fabrication of SMA-Integrated 3D Woven Corners}
Figure \ref{fig:Fab}(a) shows the fabrication of an active woven corner. We first have all the materials ready including three groups of ribbons, twenty-seven split pins, two Flexinol\textsuperscript{\textregistered} (Dynalloy Inc., USA) coiled nickel-titanium SMAs, and four copper wires (Fig. \ref{fig:Fab}(a, \textit{i})). The ribbons are laser cut from a Mylar\textsuperscript{\textregistered} polyester sheet that has a thickness 0.1905 mm. We attach adhesive Kapton\textsuperscript{\textregistered} polyimide thin films (thickness of 0.0254 mm) on both sides of the Mylar\textsuperscript{\textregistered} sheet, because Kapton\textsuperscript{\textregistered} films can withstand high temperatures up to 400°C and thermally shield the Mylar\textsuperscript{\textregistered} ribbons. All ribbons have dimensions of 135$\times$13$\times$0.2159 mm. We interweave ribbons of Groups 1 and 2, Groups 1 and 3, and then Groups 2 and 3 into a 3D surface (Fig. \ref{fig:Fab}(a, \textit{ii}--\textit{iv})). Throughout the process, we use split pins to secure the boundaries. The resulting woven structure resembles a \textit{corner} of a plain-woven basket \cite{tu2025corner}. Then, we connect the two coiled SMAs to four pre-drilled holes on the woven corner (Fig. \ref{fig:Fab}(a, \textit{v})). The locations of holes are chosen so that the step size of the robot in locomotion is maximized, which will be discussed more later. Each SMA has a free length of 19 mm (37 active coils) and is pre-stretched to an installed length of 35.6 mm, matching the distance between the attachment holes on the woven corner surface. The SMA wire diameter and coil diameter are 0.51 mm and 3.45 mm, respectively. The actuator has a nominal activation temperature of $90\,^{\circ}\text{C}$. In the austenite phase, the SMA Young's modulus and Poisson's ratio are 70 GPa and 0.33, corresponding to a shear modulus of 26.3 GPa. We use epoxy resin to fix the four ends of the alloys onto the corner surface. Next, we tie the copper wires to the four ends of the SMAs (Fig. \ref{fig:Fab}(a, \textit{vi})), and we use silver epoxy paste to bond the copper wires to SMAs. Finally, we attach an acrylic plate to the top of the woven corner as a loading tray (Fig. \ref{fig:Fab}(b), top panel). The woven corner (10 g) can carry a load that is 25 times its self-weight without significant deformation.

When we apply a direct current of 1 A to Joule-heat the SMAs through the copper wires, the pre-stretched SMAs contract and the entire corner structure folds (Fig. \ref{fig:Fab}(b), bottom panel). Due to the asymmetry of the locations of the SMAs with respect to the woven surface in the \textit{x}-\textit{z} plane, the actuation and deformation of the corner structure is asymmetric, and thus the contact and friction between each vertex of the woven corner and the ground is asymmetric. Therefore, when the woven corner is actuated, the front vertex stays still and the two back vertices slide forward by a distance of $\Delta$; while when the corner is relaxed, the back vertices stay still and the front vertex slides forward by a distance of $\Delta$. This actuation--relaxation process results in a single step forward. By cyclically implementing the actuation--relaxation process (Fig. \ref{fig:Fab}(c)), we achieve locomotion of the woven corner along a straight line (Fig. \ref{fig:Fab}(d)). Besides locomotion, we also achieve steering of the woven corner by actuating only one of the SMAs (Fig. \ref{fig:Fab}(e)). During the steering, the woven corner both moves forward and rotates, so the resulting path is a curve in the \textit{x}-\textit{y} plane (Fig. \ref{fig:Fab}(f)). Both experimental paths match well with the simulated paths (Fig. \ref{fig:Fab}(d) and (f), bottom panels, the simulation method will be detailed later). The videos of locomotion and steering can be found in Supplementary Video S1.

\begin{figure}
\centering
\includegraphics[width=1\textwidth]{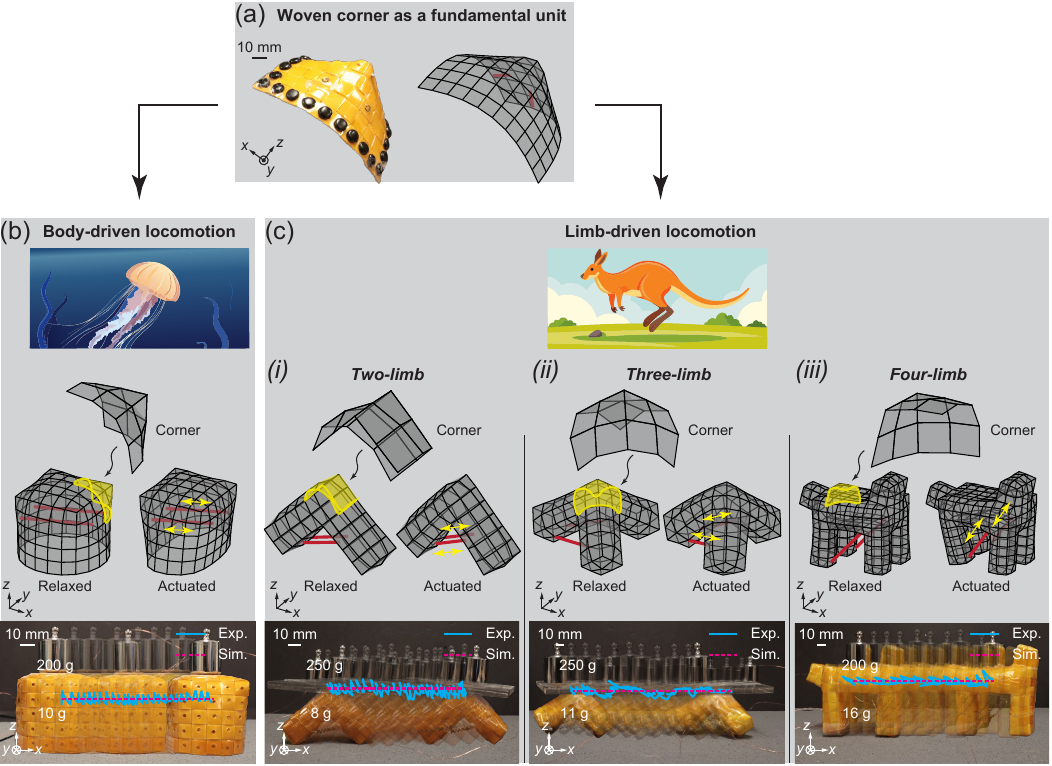}
{\caption{Corner-based modular design of 3D load-bearing woven robots. (a) Woven corner as a fundamental unit for modular assembly of 3D woven robots; (b) and (c) show two types of 3D woven robots assembled from woven corner units that utilize the mechanisms of body-driven locomotion and limb-driven locomotion, respectively. In each column in (b), (c, \textit{i}), (c, \textit{ii}), and (c, \textit{iii}), we show the simulated relaxed (unactuated) and actuated configurations of each woven robotic structure where a corner unit is highlighted; below the simulations, we show the images of each moving woven robot that are juxtaposed with the simulated and experimental locomotion paths (blue and magenta curves), and the images are ten snapshots with the same time duration between each one.}\label{fig:Modular}}
\end{figure}

\subsection*{Corner-Based Modular Design of Active 3D Weaves}
The woven corner can serve as a fundamental unit in modular design of general 3D robotic weaves that have more complex geometries and topologies. Figure \ref{fig:Modular} shows four different designs and prototypes of robotic weaves where each structure contains woven corners in their geometry. The \textit{woven cube} in Fig. \ref{fig:Modular}(b) exhibits a \textit{body-driven locomotion} where the structure moves forward through cyclic shrinking and expansion of the body. The two-legged \textit{woven walker}, three-legged \textit{woven crawler}, and four-legged \textit{woven dog} in Fig. \ref{fig:Modular}(c) all exhibit a \textit{limb-driven locomotion} where the structures move forward through cyclic bending and relaxation of the legs. Similar to the individual corner unit in Fig. \ref{fig:Fab}, all these modularly assembled woven robots are stiff enough axially to carry loads that are around 20 times their self-weight in locomotion, while flexible enough in lateral bending to enable efficient actuation. All experimental locomotion paths match well with the simulated paths.

\section*{Mechanics Modeling of 3D Woven Robots}\label{Sec:Modeling}
\begin{figure}
\centering
\includegraphics[width=1\textwidth]{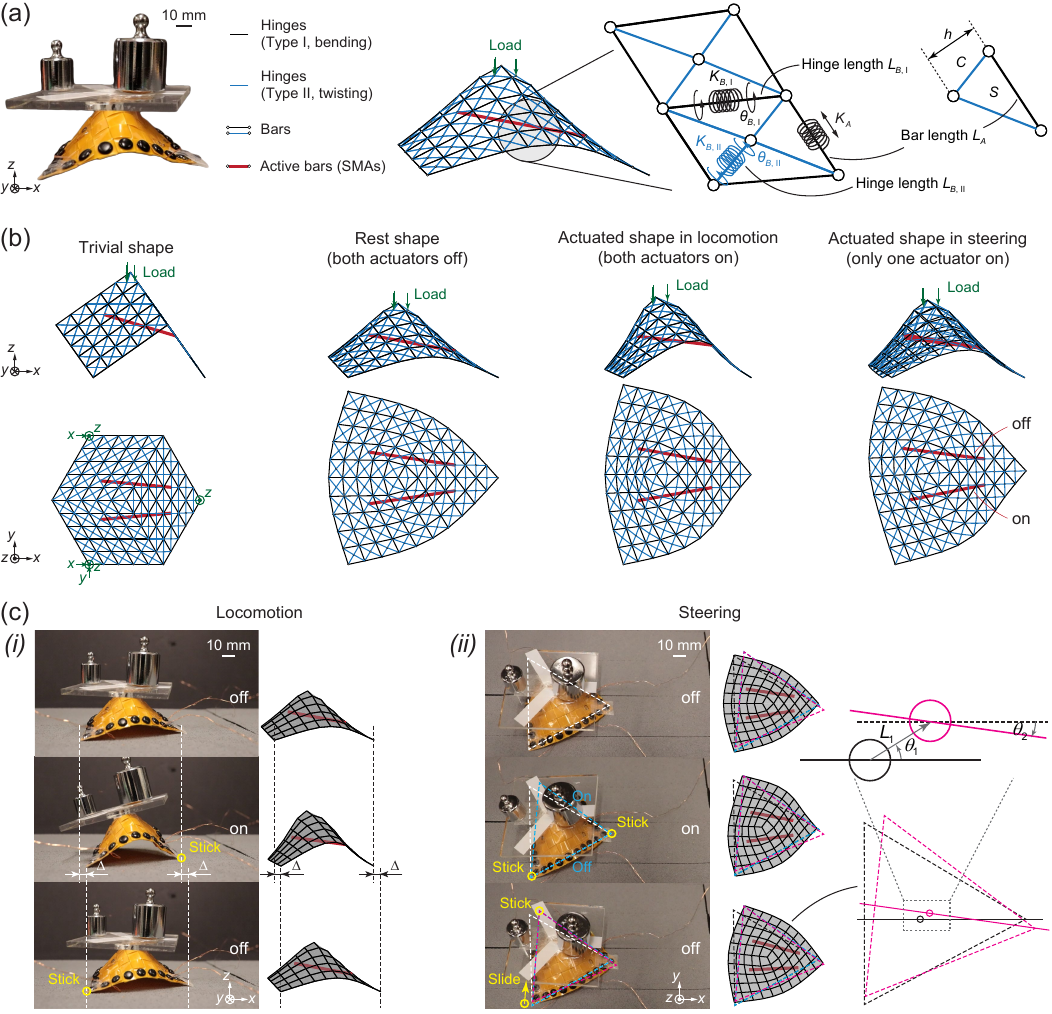}
{\caption{Mechanics modeling of 3D woven robots. (a) The bar \& hinge mesh of an active woven corner under external load. A schematic of bar and hinge elements are shown on the right; (b) The four configurations of the loaded woven corner from the bar \& hinge simulation: the trivial shape, the rest shape with both actuators off, the actuated shape in locomotion with both actuators on, and the actuated shape in steering with only one actuator on; in the trivial shape, the external load and boundary condition are shown in green; (c) (\textit{i}) and (\textit{ii}) show how we calculate the step size of locomotion and steering of the woven corner robot using the bar \& hinge simulation; the left and right columns in (\textit{i}) and (\textit{ii}) show a comparison between experimental and simulated configurations in one actuation cycle. For locomotion along a stright line (c, \textit{i}), we use $\Delta$ to represent the translational step size in one actuation cycle; while for steering along a curved line (c, \textit{ii}), we use triangles formed by the three vertices to characterize the configurations of the woven corner (the white, blue, and magenta triangles in (c, \textit{ii})); we use the magnitude $L_1$ and the counterclockwise angle from the axis of symmetry $\theta_1$ to represent the translational vector of the center of the triangle in the $x$--$y$ plane, and we use $\theta_2$ to represent the angle of rotation of the axis of symmetry of the triangle in each actuation cycle.}\label{fig:Mechanics}}
\end{figure}

We present a mechanics model of 3D woven robots based on the \textit{bar \& hinge} discretization, which is an order-reducing meshing strategy for thin-shell structures \cite{filipov2017bar}. As Fig. \ref{fig:Mechanics}(a) shows, the host structure of a woven robot is discretized by bar elements along the sides of 3D grid squares formed by interwoven ribbons (ribbon stretching) and the diagonals within in each square (ribbon shearing) \cite{tu2025corner}. A torsional spring element is attached to each bar element to characterize the ribbon bending and twisting. The stiffness of bar ($K_A$) \& hinge elements ($K_{B, \ \mathrm{I}}$, $K_{B, \ \mathrm{II}}$) are determined by material properties and mesh geometries of the woven structure \cite{tu2025corner,tu2025engineering} (Fig. \ref{fig:Mechanics}(a)). Then, to model the embedded SMA actuators, we use an \textit{active bar} element that has a stress-free length $L_A^{\mathrm{SF}}$. When the SMA is cold (unactuated), the $L_A^{\mathrm{SF}}$ is equal to the initial length in the mesh, so no actuation force will be generated; when the SMA is heated (actuated), the $L_A^{\mathrm{SF}}$ is set to the original length of unstretched cold SMAs, so a self-retracting force will be generated in the bar to actuate the structure. The effective stiffness of active bar elements is $K_A = Gd^4/8nD^3$ where $G$ is the shear modulus of the SMA wire in austenite phase, $d$ is the wire diameter, $D$ is the coil diameter, and $n$ is the number of coils \cite{Shigley2014ME}.

Based on the bar \& hinge mesh, we find the actuated and unactuated shapes of a woven robot using nonlinear mechanics simulation. We start with the trivial shape of a woven robot, for instance, for the woven corner, the trivial shape is a non-smooth corner of a trirectangular tetrahedron (Fig. \ref{fig:Mechanics}(b)). Next, we assign $\pi$ to stress-free angles of all hinge elements $\theta_B^{\mathrm{SF}}$; for inactive bar elements and unactuated active bar elements (SMAs), we calculate the initial lengths in the trivial shape as stress-free lengths $L_A^{\mathrm{SF}}$; for actuated active bar elements (SMAs), we assign the original lengths of unstretched cold SMAs to $L_A^{\mathrm{SF}}$. The internal stress of $M$ bent hinges and $N$ stretched bars, and the external load result in a non-zero potential energy $\Pi = \sum\limits_{m = 1}^M {\frac{1}{2}{K_{B,m}}{{\left( {{\theta _{B,m}} - \theta_{B,m}^{\mathrm{SF}}} \right)}^2}}  + \sum\limits_{n = 1}^N {\frac{1}{2}{K_{A,n}}{{\left( {{L_{A,n}} - L_{A,n}^{\mathrm{SF}}} \right)}^2}} + {{\bf{f}}^T}{\bf{u}}$, where $\theta_{B,m}$ and $L_{A,n}$ indicate the current hinge angles and bar lengths, $\theta_{B,m}^{\mathrm{SF}}$ and $L_{A,n}^{\mathrm{SF}}$ indicate the stress-free states, and $\bf{f}$ and $\bf{u}$ are the nodal load and nodal displacement vectors.

We use an incremental Newton-Raphson solver to minimize the energy $\Pi$ where we use a minimal boundary condition to eliminate six rigid body degrees of freedom. The resulting rest shape (with both actuators off), actuated shape in locomotion (with both actuators on), and actuated shape in steering (with only one actuator on) of the woven corner robot are shown in Fig. \ref{fig:Mechanics}(b). Using these simulated shapes, we can calculate the step size of woven robots. In locomotion, we use a linear step size $\Delta$ to characterize the `stick-slip' movement \cite{wang2017design,farhadi2025origami,rafsanjani2018kirigami}, which is the offset distance between two consecutive relaxed shapes in the $x$--$z$ plane (Fig. \ref{fig:Mechanics}(c, \textit{i})). In steering, we use a translational vector, represented by a magnitude $L_1$ and a rotational angle $\theta_1$ (counterclockwise starting from the current centerline of the structure), to characterize the linear motion of the center of the robot; we use another rotational angle $\theta_2$ (clockwise) to characterize the rotation of the centerline of the robot. The three parameters $L_1$, $\theta_1$, and $\theta_2$ together capture the translation and rotation of the woven robot in the $x$--$y$ plane in a single steering step. Continuous simulation and plotting of each step generates complete locomotion/steering paths in Fig. \ref{fig:Fab} and \ref{fig:Modular}. The experimental paths match with the results from our reduced-order model. After we obtain the rest shapes, we can extract the linear stiffness matrix of the structure for eigenmode analysis. These eigenmodes can be used to systematically determine the loading bearing and actuation mechanisms of a woven robot, which is be demonstrated next. The simulation code used in this study will be openly available on GitHub \cite{Tu2026BarHinge}.

\section*{Locomotion, Load Bearing, and Damage Resilience of 3D Woven Robots}\label{Sec:Scaling}
Next, using the mechanics-based simulation and experiments, we quantitatively characterize the mode-dependent stiffness, load bearing capacity during locomotion, and resilience to extreme deformations of 3D woven robots.

\subsection*{Stiff mode-enabled load bearing and flexible mode-enabled locomotion}
\begin{figure}
\centering
\includegraphics[width=1\textwidth]{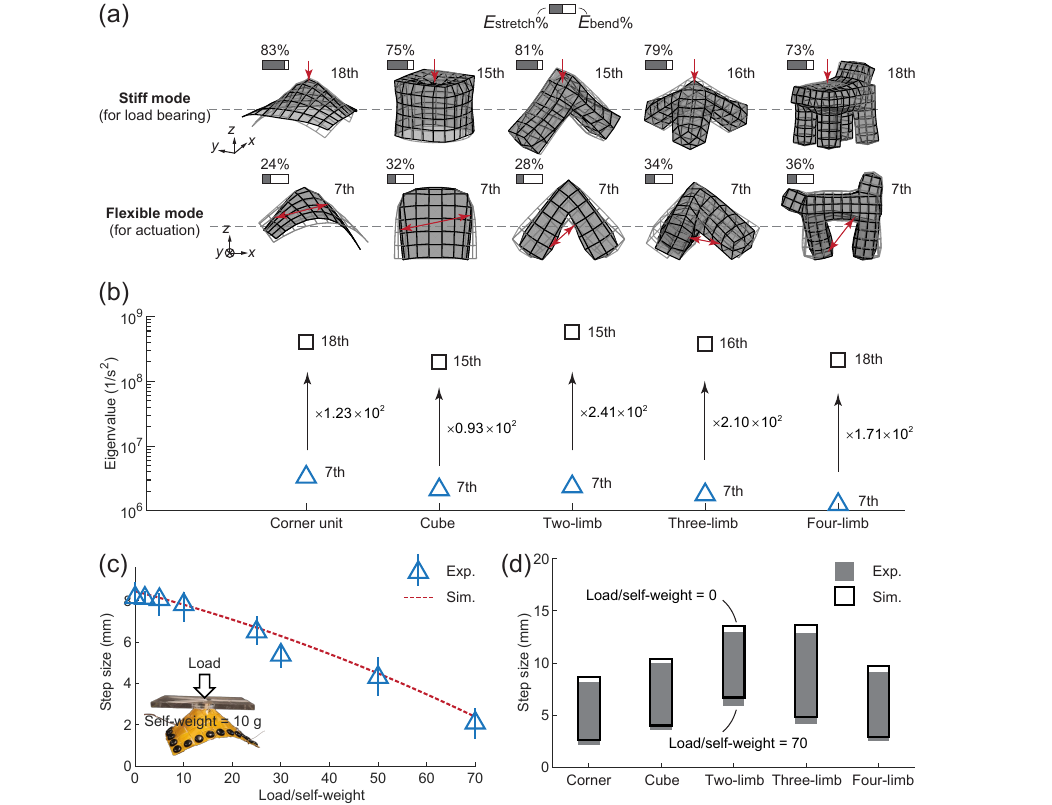}
{\caption{Locomotion and load bearing capacity of 3D woven robots. (a) Finding stiff and flexible modes of our 3D woven robotic structures using boundary condition-free eigenmode analysis. The flexible modes are used to decide where to install the SMA actuators for efficient actuation (red double arrows). The stiff modes have the largest deformation along the $z$ direction among all high-order modes beyond the 15th; we use these stiff modes to decide where to put the external load (red arrows). For each eigenmode, we show the strain energy ratio of the deformation mode where the shaded area represents the stretching/compression energy of bars while the blank area represents the bending/twisting energy of hinges in our bar \& hinge model; (b) The eigenvalues of the flexible and stiff modes in (a) with the corresponding ratio between the two; (c) The locomotion step size of the woven corner robot changes as the amount of external load increases. The error bars are calculated based on three tests of three samples fabricated the same way; (d) The range of variation of the locomotion step size of each 3D woven robot as we change the external load from 0 to 70 times their self-weights.}\label{fig:Stiffness}}
\end{figure}

For a newly designed 3D woven structure, we can systematically find the most stiff and most flexible deformation modes using linear eigenvalue analysis. Figure \ref{fig:Stiffness}(a) shows the stiff and flexible modes of each robotic structure. The 7th eigenmodes represent the most flexible mode with elastic deformation, while the first six modes represent free body movement and are irrelevant to the robotic function. In the elastic flexible modes, the woven ribbons are mainly engaged in lateral bending (low stiffness deformation) instead of axial stretching/compression (high stiffness deformation), which is characterized by the stretching--bending energy ratios of each eigenmode given in Fig. \ref{fig:Stiffness}(a). We use these flexible modes to decide where to install the SMA actuators for efficient actuation (indicated by red double arrows in the lower panels in Fig. \ref{fig:Stiffness}(a)). While the locations of SMAs are not unique, we pick two points that experience a large deformation and can maximize the step size in locomotion.

Then, we find the stiff modes by picking the eigenmodes that have the largest deformation along the $z$ direction (the direction of gravity) among all high-order modes beyond the 15th, as a significant eigenvalue gap from the 7th emerges after the 15th. In these stiff modes, the woven ribbons are mainly engaged in axial stretching/compression instead of lateral bending. We use these stiff modes to decide where to place the external load (indicated by red arrows in the upper panels in Fig. \ref{fig:Stiffness}(a)), which allows us to efficiently load the woven robots axially. Between the eigenvalues of flexible and stiff modes, a large gap exists where the eigenvalues of stiff modes are more than 100 times the eigenvalues of flexible modes (Fig. \ref{fig:Stiffness}(b)). This large eigenvalue gap indicates the high deformation energy needed for stiff modes and low energy needed for flexible modes, which further justifies our methodology to decide the locations for actuators and external load \cite{tu2024origami, filipov2015origami}.

As the external load increases from 0 to 70 times the self-weight, the woven corner robot maintains the ability of locomotion with a slow decrease in the step size, which is verified by both simulations and experiments (Fig. \ref{fig:Stiffness}(c)). This robustness of the locomotion capacity regarding the amount of external load can be observed for all woven robots we designed. As Fig. \ref{fig:Stiffness}(d) shows, when the external load increases from 0 to 70 times the self-weight of each robot, the step size reduces by 2/3 overall.

\begin{figure}
\centering
\includegraphics[width=1\textwidth]{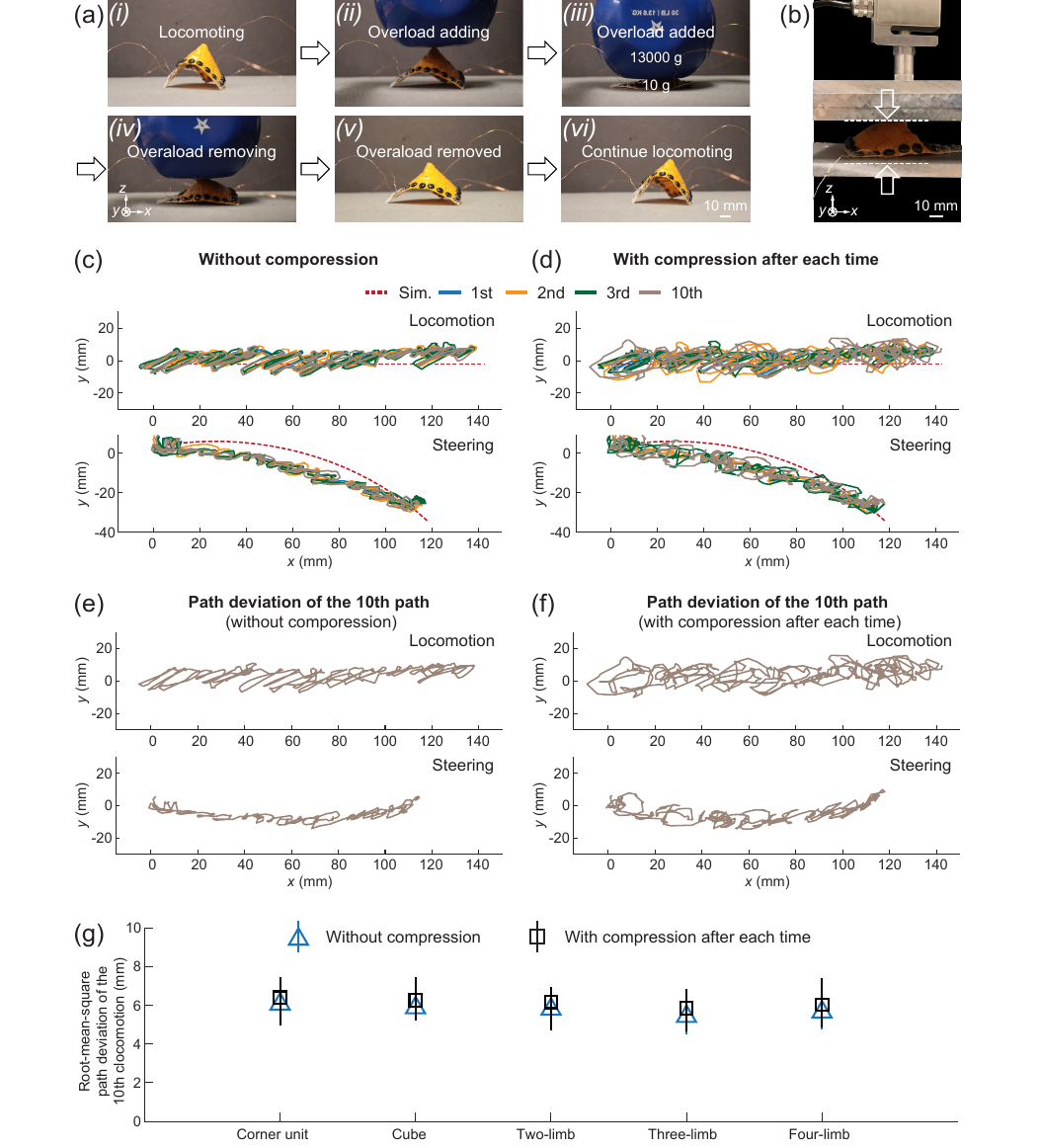}
{\caption{Damage resilience of 3D woven robots. (a) Snapshots of a video where a woven corner robot recovers its initial geometry and continues locomoting after an extreme compression from the external load 1300 times its self-weight. The robot can withstand this extreme compression multiple times and resume locomotion, which is shown in Supplementary Video 2; (b) The experimental setup to compress a woven robotic structure to a height that is 1/4 of its initial height; (c) The consecutively repeated locomotion and steering paths of the woven corner robot without compression; (d) The consecutively repeated locomotion and steering paths of the woven corner robot with compression after each path; the compression is done using the setup in (b); (e) and (f) give the path deviation between the experimental path and the simulated path in the 10th locomotion in (c) and (d); (g) the root-mean-square path deviation of the 10th locomotion of each woven robot without compression and with compression after each compression cycle. The error bars are calculated based on three tests of three samples fabricated the same way, where the upper boundary is the highest value and the lower boundary is the lowest value among all three tests.}\label{fig:Resilience}}
\end{figure}

\subsection*{High resilience to extreme deformations}
In addition to the excellent load bearing capacity, our woven robots exhibit extraordinary resilience when subjected to extreme deformations. Figure \ref{fig:Resilience}(a) shows that the woven corner robot recovers its initial geometry and continues locomotion after it is crushed into a flat state by an external load that is 1300 times its self-weight. The robot can withstand this extreme compression multiple times and resume locomotion, as shown in Supplementary Video S2.

We further quantitatively characterize the mechanical resilience of the woven corner robot. First, we repeat the locomotion and show that the robot follows a similar path every time despite some variation in local steps (Fig. \ref{fig:Resilience}(c)). Then, we explore the same repeatability in locomotion but now with an extreme load applied after every locomotion cycle. As Fig. \ref{fig:Resilience}(d) shows, the 1st locomotion cycle is before any extreme load was applied, then for the 10th locomotion cycle the extreme load was applied 9 times (once after every locomotion cycle). For the repeatability of the testing, we use a Mark-10 universal testing machine to consistently compress the woven corner to a height that is 1/4 of the initial height (Fig. \ref{fig:Resilience}(b)). We also give the path deviation of the 10th experimental locomotion cycle from the simulated path in both cases, as shown in Fig. \ref{fig:Resilience}(e) and (f).

Compared to the paths without extreme compression, the paths with compression after each locomotion cycle exhibit slightly larger fluctuations and real-time path deviations, but the overall trends and deviations at the final positions are similar. Therefore, loading cycles of extreme compressions have little impact on the overall motion of the woven corner robot, which is demonstrated through the small difference between the root-mean-square path deviation of the 10th locomotion of the robot without compression and the deviation with compression after each time (Fig. \ref{fig:Resilience}(g)). This excellent mechanical resilience can be observed for all woven robots presented in this paper (Fig. \ref{fig:Resilience}(g)).

The resilience of woven robotic systems comes from two aspects: when 3D woven structures go through extreme deformations, the constituent woven ribbons exhibit segmental elastic buckling instead of buckling with plastic deformations, because the filaments are not rigidly tied to each other in weaving \cite{krankel2026stiffness,tu2025corner}. Furthermore, when the embedded SMAs are cold and in a plastic martensite phase, they can withstand large arbitrary deformations. Once the imposed deformation is removed and the SMAs are heated again, they will recover their austenite phase and continue to contract and actuate the woven structure as before \cite{baxevanis2014fracture,baxevanis2015fracture}. The resilience of host woven structures and SMA actuators lead to the system-level resilience of the 3D woven robots.

\section*{Conclusions}\label{Sec:Conclusion}
In this work, we presented a general framework for integrating coiled SMAs into three-dimensional plain-woven structures to create load-bearing robotic textiles. Inspired by basket weaving, we defined the woven corner as a modular unit that combines high axial stiffness for load bearing with low bending stiffness for actuation. By assembling corner units into different topologies, we realized body-driven and limb-driven woven robots that locomote and steer under simple electrothermal input while carrying loads exceeding 70 times their self-weight.

We present a reduced-order bar \& hinge model that captures the mechanics of these systems through compliant hinge elements and self-contracting active bar elements (the simulation code will be openly available on GitHub \cite{Tu2026BarHinge}). The model predicts step sizes and steering angles, and it reveals a clear stiffness anisotropy between filament-stretching-dominated and filament-bending-dominated eigenmodes. This anisotropy leads to a practical design strategy for 3D woven robots: stiff high-order modes identify favorable loading directions, whereas flexible low-order modes indicate efficient actuator placement. Consistent with this mechanism, simulations and experiments show that locomotion remains robust over a broad range of external loads (0--70 times the self-weight), even though the step size gradually decreases as the load increases.

Another key result is the exceptional resilience of these active 3D weaves under extreme deformation. The woven architecture accommodates overload through elastic buckling of constituent filaments, while the SMA actuators can undergo large deformation in the martensitic state and recover their actuation function afterward. As a result, the entire woven robot system preserves geometry and repeatable motion even after repeated severe compression.

Overall, this study establishes active 3D weaves as a promising platform for robotic textiles that unite load bearing, programmable locomotion, and damage tolerance in a single system. These results open opportunities for wearable machines, assistive devices, rescue robots, and other multifunctional soft robotic systems, and they provide a foundation for future work on improved control, larger-scale assemblies, and alternative active materials.


\section*{Acknowledgements}
The authors acknowledge support from the Air Force Office of Scientific Research under award number FA9550-22-1-0321. The paper reflects the views and opinions of the authors, and not necessarily those of the funding entities.

\bibliography{sample}

\end{document}